\documentclass[letterpaper, 10 pt, journal, twoside]{IEEEtran}

\IEEEoverridecommandlockouts                              

\usepackage{graphics} 
\usepackage{epsfig} 
\usepackage{mathptmx} 
\usepackage{times} 
\usepackage{amsmath} 
\usepackage{amssymb}  
\usepackage{amsfonts}
\usepackage{hyperref}

\usepackage{subcaption} 
\usepackage{multirow}
\usepackage{tabularx}
\usepackage{makecell}
\usepackage{booktabs}
\usepackage{array}
\usepackage{cite}

\usepackage{graphicx}
\usepackage{caption}
\usepackage{subcaption}

\DeclareMathAlphabet{\mathcal}{OMS}{cmsy}{m}{n}

\makeatletter
\usepackage{xspace}
\DeclareRobustCommand\onedot{\futurelet\@let@token\@onedot}
\def\@onedot{\ifx\@let@token.\else.\null\fi\xspace}

\makeatother

\title{GMFlow: Global Motion-Guided Recurrent Flow for 6D Object Pose Estimation
}

\author{
Xin Liu$^{1}$, 
Shibei Xue$^{1}$,  
Dezong Zhao$^{2}$, Shan Ma$^{3}$, and Min Jiang$^{4}$
\thanks{This work was supported in part by the National Natural Science Foundation of China under Grant 62273226 and Grant 61873162. (Corresponding author: Shibei Xue)}%
\thanks{$^{1}$Xin Liu and Shibei Xue are with Department of Automation, Shanghai Jiao Tong University, Shanghai, P. R. China.
 {\tt\small liu.xin@sjtu.edu.cn, shbxue@sjtu.edu.cn}}%
\thanks{$^{2}$Dezong Zhao is with James Watt School of Engineering, University of Glasgow, Glasgow, United Kingdom.
{\tt\small Dezong.Zhao@glasgow.ac.uk}}%
\thanks{$^{3}$Shan Ma is with School of Automation, Central South University, Changsha, P. R. China.
{\tt\small shanma@csu.edu.cn}}
\thanks{$^{4}$Min Jiang is with School of Electronics and Information Engineering, Soochow University, Suzhou, P. R. China.
{\tt\small jiangmin08@suda.edu.cn}
}}

\begin{document}

\maketitle


\markboth{}{Liu \MakeLowercase{\textit{et al.}}: GMFlow: Global Motion-Guided Recurrent Flow for 6D Object Pose Estimation}

\begin{abstract}
6D object pose estimation is crucial for robotic perception and precise manipulation. Occlusion and incomplete object visibility are common challenges in this task, but existing pose refinement methods often struggle to handle these issues effectively. To tackle this problem, we propose a global motion-guided recurrent flow estimation method called GMFlow for pose estimation.
GMFlow overcomes local ambiguities caused by occlusion or missing parts by seeking global explanations. We leverage the object's structural information to extend the motion of visible parts of the rigid body to its invisible regions. Specifically, we capture global contextual information through a linear attention mechanism and guide local motion information to generate global motion estimates.
Furthermore, we introduce object shape constraints in the  flow iteration process, making flow estimation suitable for pose estimation scenarios.
Experiments on the LM-O and YCB-V datasets demonstrate that our method outperforms existing techniques in accuracy while maintaining competitive computational efficiency.
\end{abstract}


\section{INTRODUCTION}
6D pose estimation is a fundamental task in robotics and computer vision that extracts the position and orientation of an object in three-dimensional space from a single image. This technology has far-reaching applications across various domains of robotics, including  object manipulation, industrial automation,  and self-driving vehicles.  Therefore, exploring how to provide precise object  pose information helps enhance robots' environmental perception and operational accuracy, driving advancements and innovations in robotics technology.

The development of deep learning has greatly advanced 6D pose estimation technology. 
\begin{figure}[t]
    \centering
    \begin{subfigure}[t]{0.49\linewidth}
        \centering
        \includegraphics[width=\linewidth]{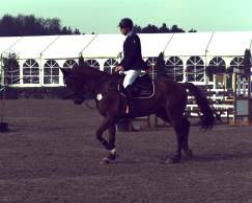}
        \caption{Frame 1}
        \label{fig:image1}
    \end{subfigure}
    \hfill
    \begin{subfigure}[t]{0.49\linewidth}
        \centering
        \includegraphics[width=\linewidth]{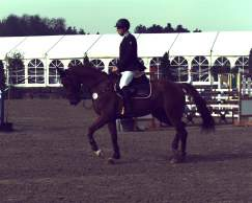}
        \caption{Frame 2}
        \label{fig:image2}
    \end{subfigure}

    \vspace{0.1cm} 

    \begin{subfigure}[t]{0.49\linewidth}
        \centering
        \includegraphics[width=\linewidth]{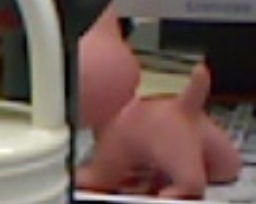}
        \caption{Occluded Object}
        \label{fig:image3}
    \end{subfigure}
    \hfill
    \begin{subfigure}[t]{0.49\linewidth}
        \centering
        \includegraphics[width=\linewidth]{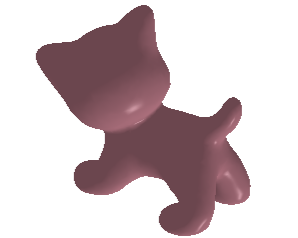}
        \caption{Cat's rendered image}
        \label{fig:image4}
    \end{subfigure}

    \vspace{0.1cm} 

    \begin{subfigure}[t]{0.49\linewidth}
        \centering
        \includegraphics[width=\linewidth]{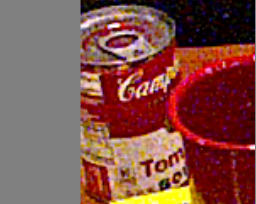}
        \caption{Partially missing object}
        \label{fig:image5}
    \end{subfigure}
    \hfill
    \begin{subfigure}[t]{0.49\linewidth}
        \centering
        \includegraphics[width=\linewidth]{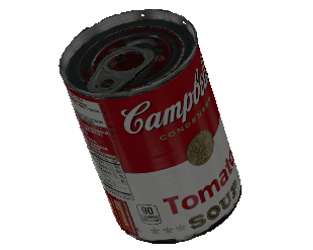}
        \caption{Can's rendered image}
        \label{fig:image6}
    \end{subfigure}

    \caption{\textbf{Challenges of flow-based method in 6D pose estimation task.} Flow is commonly used to estimate the pixel motion vector field between two frames in an image sequence. Examples are shown in (a) and (b)\cite{janai2017slow}. The objects in the images are not necessarily rigid bodies, and their changes are typically limited in magnitude. However, pose estimation tasks differ in this regard. Although rendered images, such as (d) and (f), are complete, objects in the target image may be occluded or incomplete, as in (c)\cite{krull2015learning} and (e)\cite{xiangposecnn}.}
    \label{fig: intro}
\end{figure}
Among learning-based methods, multi-stage strategies have shown the best performance\cite{sundermeyer2023bop}. These methods typically first localize target instances in an image through 2D object detection or segmentation, then predict the object pose corresponding to the 2D region or mask. In the final step, pose refinement techniques are applied, with some methods using the Iterative Closest Point (ICP) algorithm to further enhance accuracy\cite{besl1992method,sundermeyer2023bop,xiangposecnn}. However, the ICP algorithm relies on depth information, which would be costly or difficult to obtain in practical applications. Therefore, pose refinement methods based solely on RGB images  have emerged\cite{li2018deepim,labbe2020cosypose}. These methods adopt a render-and-compare concept: they use an initial estimated pose to render an image, compare the rendered result with the original one, and refine the pose  iteratively. Although these methods perform excellently with the support of large-scale training data, they still face challenges in practical applications. In particular, obtaining a large amount of high-quality pose labels is often very difficult, which limits the training process.

To overcome this limitation, Hu et al.\cite{hu2022perspective}  introduce  flow networks, e.g. RAFT\cite{teed2020raft}, to predict 2D  flow from a rendered image to a real input image, and use a Perspective-n-Point (PnP) algorithm to calculate the pose. To further optimize this method, Hai et al.\cite{hai2023shape}  use 2D projections of the object's 3D points to compute  flow, and employ a neural network to predict the pose from the flow, achieving a fully differentiable pose refinement process.
However, these methods still simply apply the RAFT framework for  flow processing, overlooking the unique characteristics of pose estimation scenarios. We can render complete images since the full 3D point information is available\cite{ravi2020pytorch3d}. But target images would be incomplete in real-world scenarios (as shown in Fig. \ref{fig: intro}). This incompleteness occurs for two main reasons: first, portions of the object may be occluded by other objects, and second, parts of the object may be missing due to image cropping. This poses a challenge for existing  flow-based pose refinement methods, as convolutional neural networks struggle to estimate the 2D motion of points that are only visible in the rendered image but are missing in the real image.

To address this issue, we propose GMFlow, a \textbf{G}lobal \textbf{M}otion-Guided Recurrent \textbf{F}low for 6D Object Pose Estimation. Our strategy still adheres to the render-and-compare framework, which renders images from initial poses and compares them with real images to compute flow for pose estimation. Conceptually, we draw inspiration from the idea of Geoffrey Hinton in 1976, which resolves local ambiguities through global explanations\cite{hinton1976using}. Under the rigid body assumption, the motion of different points on an object is similar and uniform. This principle also applies to the 2D motion between rendered and real images. Therefore, we can transfer motion information from visible parts to invisible parts. To achieve this, we use an attention mechanism to aggregate global contextual information of the object and combine it with local motion information to generate global motion estimates\cite{vaswani2017attention,jiang2021learning,shaker2023swiftformer}. This global motion information is then fed into a Gated Recurrent Unit (GRU) to iteratively estimate the flow between the rendered and real images\cite{cho2014properties}, aiding in pose calculation. Furthermore, we incorporate the object's 3D shape into the recurrent flow estimation process through projection transformations, forming a closed-loop iteration between the flow and pose. This enhances the applicability of flow estimation in 6D pose tasks.

Through these designs, 
GMFlow effectively handles incomplete object visibility, achieving state-of-the-art accuracy and robustness on benchmark datasets while maintaining competitive runtime.

\section{RELATED WORKS}
\subsection{Object Pose Estimation}
The field of pose estimation has made significant strides, largely due to advancements in deep learning\cite{sundermeyer2023bop,vaswani2017attention}. Early approaches directly regressed 6D object poses from input images\cite{wang2021gdr,xiangposecnn,jantos2023poet}. While these methods are end-to-end trainable and computationally efficient, they often fall short in terms of accuracy. To address this limitation, researchers began exploring indirect methods for pose prediction.
Two prominent indirect approaches emerge, including keypoint-based methods\cite{peng2019pvnet,guan2021high} and dense correspondence-based methods\cite{shugurov2021dpodv2,haugaard2022surfemb}. The former detects a small number of semantically meaningful feature points, such as object corners or edges, and then solves for the pose using the Perspective-n-Point (PnP) algorithm\cite{lepetit2009ep}. This approach offers computational efficiency but would be sensitive to occlusions. The latter establishes numerous uniformly distributed correspondence points on the object surface. Although computationally more intensive, this method demonstrates greater robustness to partial occlusions and often yields higher accuracy.
Recent research has further refined these methods, introducing differentiable PnP algorithms to enhance performance\cite{chen2022epro}. However, the practical accuracy of these approaches still leaves room for improvement. Therefore, our work applies pose refinement to achieve more precise results.

\begin{figure*}[ht]
    \centering
    \includegraphics[width=\textwidth]{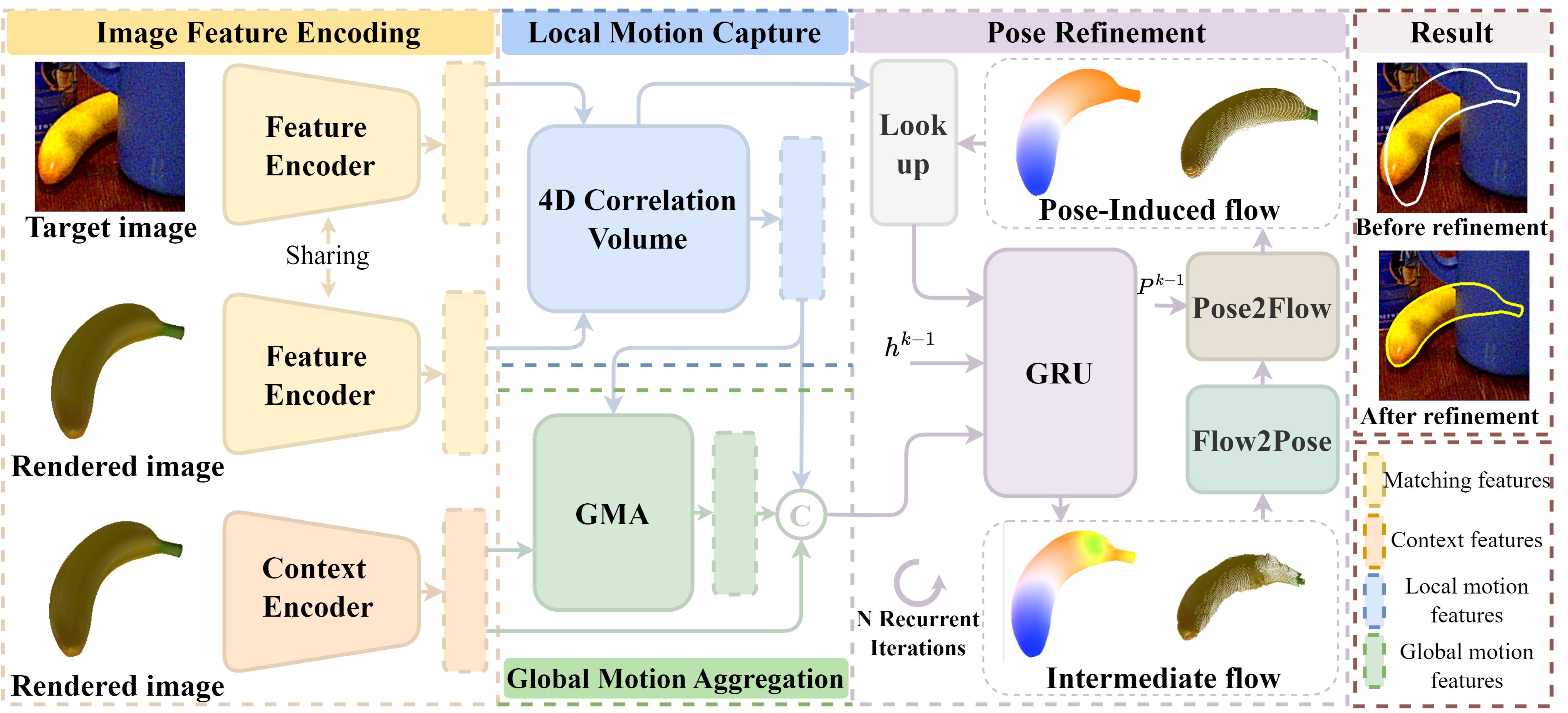}
    \caption{\textbf{Overview of the proposed method.} }
    \label{fig: overview}
\end{figure*}
\subsection{Object Pose Refinement}
Traditional pose refinement methods typically rely on depth information, with the Iterative Closest Point  (ICP) algorithm being the most representative\cite{besl1992method,xiangposecnn}. The ICP algorithm iteratively minimizes the distance between corresponding points in two point clouds to determine an optimal rigid transformation (rotation and translation). While this method is highly interpretable, its application is limited due to the difficulty of obtaining depth images in many scenarios.
In recent years, some studies have adopted a render-and-compare strategy for pose refinement\cite{li2018deepim,hu2022perspective,xu2022rnnpose,hai2023shape}. This approach first renders a synthetic image using the object's 3D model and the current estimated pose, then compares it with an actual input image. By adjusting pose parameters to maximize the similarity between the rendered and actual images, these methods can optimize directly in the image space without explicit feature matching. However, existing methods often simply employ  flow estimation techniques represented by RAFT\cite{teed2020raft} in comparison process, neglecting  shape priors and geometric context of objects in 6D pose estimation tasks. 
In contrast, our method takes full advantage of the object's rigid body properties and leverages shape information to constrain the flow.  These improvements not only enhance the accuracy of pose refinement but also accelerate convergence speed.

\section{METHODS}
Given a cropped RGB image $I_{t}$, the initial pose $\mathbf{P}_{0}$ and the 3D model $\mathcal{M}$ of the target object, our goal is to obtain an accurate 6D pose $\mathbf{P}=\left[ \mathbf{R}|\mathbf{t} \right]$. Here, $\mathbf{R}\in SO\left( 3 \right)$ denotes a rotation matrix of the object, and $\mathbf{t}\in \mathbb{R} ^3$ represents a translation vector.

Fig. \ref{fig: overview} provides a comprehensive overview of our method's architecture. Our GMFlow framework is built upon four fundamental components: image feature extraction, local motion capture, global motion aggregation, and pose refinement. The features from the target and rendered images are first encoded and then associated through local motion capture and global motion aggregation. During the pose refinement stage, flow, correlation, aggregated information, and the hidden state $h^{k-1}$ are simultaneously input into the GRU\cite{cho2014properties}. The GRU initially predicts an intermediate  flow and subsequently generates pose correction. By combining the correction with the pose from the previous iteration $p^{k-1}$, we generate a pose-induced flow used for querying corresponding features. After $N$ iterations, we obtain the final pose. 
In the following, we will discuss each of these components in detail.

\subsection{Image Feature Encoding}
\label{sec: extractor}
\textbf{Feature Encoding.}
We employ a CNN network with shared weights to encode features from both the target image $I_{t}\in \mathbb{R} ^{3\times H\times W}$ and rendered image $I_{r}\in \mathbb{R} ^{3\times H\times W}$, resulting in matching features at 1/8 resolution $\{ \mathcal{F}_t,\mathcal{F}_r\in \mathbb{R} ^{D\times \frac{H}{8}\times \frac{W}{8}}\} $. The feature encoder is constructed with six residual blocks, strategically distributed across 1/2, 1/4, and 1/8 resolution levels—two blocks at each level.  These matching features, extracted from both the target and rendered images, serve as the basis for computing a 4D correlation volume $\mathcal{V} _{corr}$, which is subsequently used to generate local motion features $\mathcal{F}_{m}$. For a more comprehensive understanding of this approach, we refer readers to the detailed implementation in RAFT\cite{teed2020raft}.

\textbf{Context Encoding.} Concurrently, we employ a context encoder with a structure identical to the feature encoder to extract contextual information from the rendered image $I_{r}$. 
After obtaining the initially encoded features, we equally divide them along the channel dimension into two parts, each with 128 channels. For the first part, we apply the \textit{tanh} activation function, limiting its output to the  range (-1, 1), serving as the hidden state feature $h_{k-1}$ input to the GRU unit\cite{cho2014properties}. For the second part, we use the \textit{ReLU} activation function to ensure non-negative output, resulting in contextual features. These contextual features are then fused with local motion features, global motion features, and flows, collectively serving as input to the GRU unit.

\subsection{Global Motion Information Capture}
\label{sec: global}
Current mainstream pose estimation methods based on RAFT\cite{teed2020raft} typically introduce 4D correlation volumes to characterize the visual similarity between the rendered and target images. These methods then use a flow  to query the 4D correlation volumes, thereby obtaining 2D motion features. Through iterative optimization of these motion features, residual flow can be derived. This approach effectively estimates pixel displacements from the rendered image to the target image in scenarios without occlusion. However, in scenes with significant occlusion, the performance of these methods often falls short of expectations\cite{hu2022perspective,hai2023shape}.

\begin{figure}[t]
    \centering
    \includegraphics[width=\linewidth]{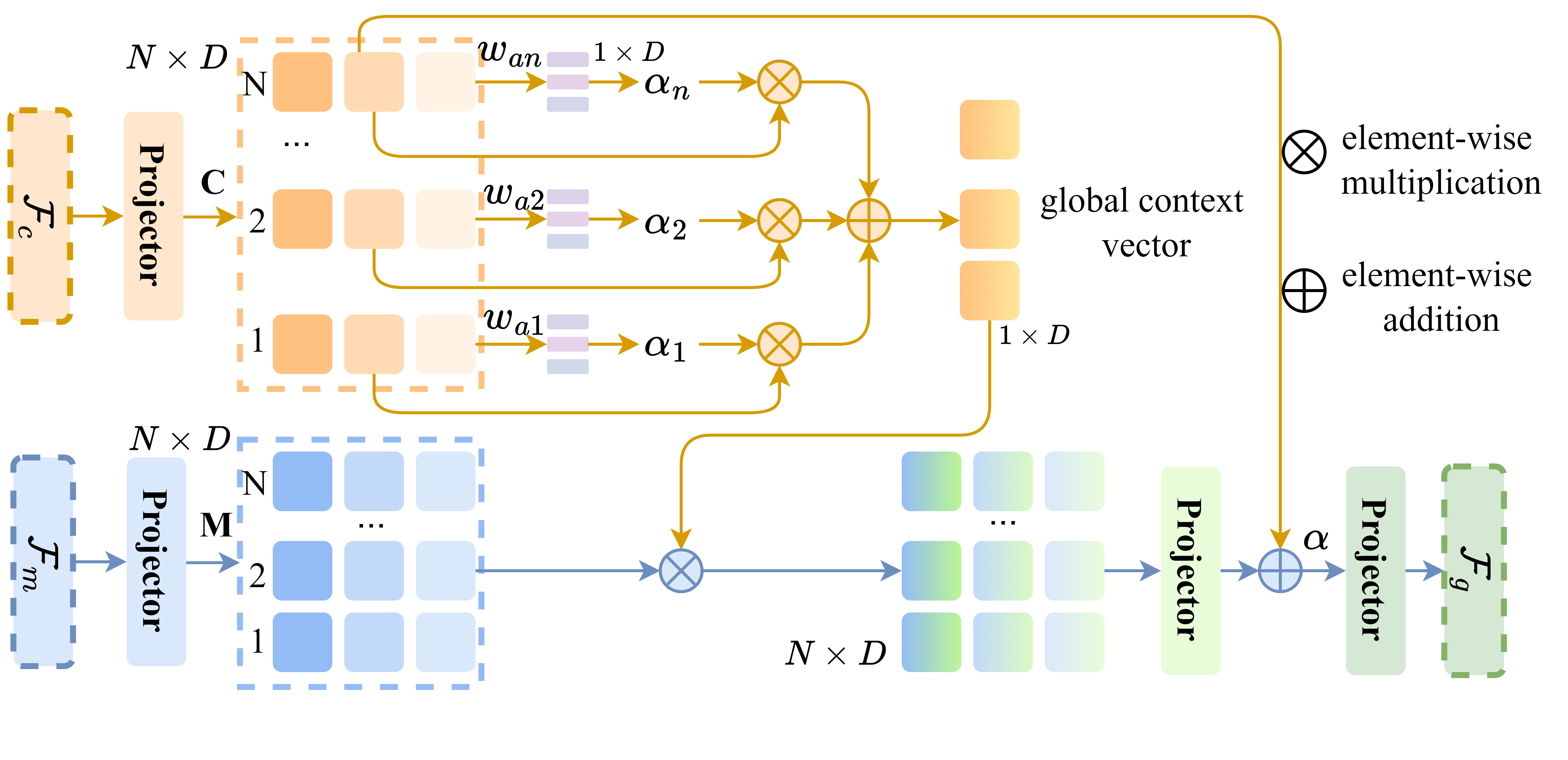}
    \caption{The structure diagram of the global motion capture module.}
    \label{fig: gmc}
\end{figure}

To address this issue, we propose a module for extracting global motion information, called the Global Motion Capture (GMC) module. Its overall architecture is illustrated in Fig. \ref{fig: gmc}. Our design is based on the following insight: local ambiguities must be resolved by finding the best global interpretation. We leverage context  features that capture the object's overall appearance to compute a global context vector. This vector then guides the extraction of 2D local motion features, resulting in a comprehensive representation of global motion features\cite{vaswani2017attention,shaker2023swiftformer}.
This approach not only maintains high accuracy in scenarios without occlusion but also provides more robust and accurate motion estimation in complex occluded scenes. By integrating global contextual information, our GMC module can better understand and predict the overall motion of objects, thus overcoming the limitations of traditional methods when dealing with local  incompleteness.

In our proposed method, we initially employ two projection functions, denoted as $\theta(\cdot)$ and $\phi(\cdot)$, to transform the context feature $\mathcal{F}_{c} \in \mathbb{R}^{D \times H \times W}$ and the motion feature $\mathcal{F}_{m} \in \mathbb{R}^{D \times H \times W}$ into $\textbf{C} \in \mathbb{R}^{N \times D}$ and $\textbf{M} \in \mathbb{R}^{N \times D}$, respectively. Here, $N=H \times W$ represents the spatial dimension, where $H$ and $W$ denote the height and width of the feature map. The parameter $D$ refers to the channel dimension of the features. Subsequently, we multiply \textbf{C} with a learnable parameter vector $\textbf{w}_\textbf{a}\in \mathbb{R} ^{D}$ and apply an appropriate scaling to generate a global attention query vector $\textbf{q}\in \mathbb{R} ^{N}$. We then apply normalization to the attention query vector $\textbf{q}$, ensuring that its sum along the spatial dimension equals unity. The above procedure can be formulated as:
\begin{equation}
    \textbf{q}={{\theta(\mathcal{F}_{c})\cdot \textbf{w}_\textbf{a}}/{\sqrt{D}}},
\end{equation}
 Subsequently, we perform a weighted summation operation on the normalized $\textbf{C}$ using $\textbf{q}$ to derive the global context vector $\textbf{g} \in \mathbb{R}^D$. This operation can be formally expressed as:
\begin{equation}
\textbf{g} = \sum_{i=1}^N \textbf{q}_i \cdot \textbf{C}_i,
\end{equation}
where $\textbf{q}_i$ and $\textbf{C}_i$ represent the $i$-th elements of $\textbf{q}$ and $\textbf{C}$ respectively.
Finally, we conduct element-wise multiplication between the global context vector $\textbf{g}$ and the matrix $\textbf{M}$, effectively capturing the intricate interactions between global and local information. This intermediate result subsequently undergoes a series of learnable transformations and residual connections, further refining and fusing the extracted features.
In the final stage of the global motion capture, we introduce a learnable scalar parameter $\alpha$, which dynamically modulates the influence of the global information. Following this, we apply a composite operation $\tau(\cdot)$, encompassing both reshaping and dimension permutation, to reconstruct the feature map to its original spatial configuration. This process yields  our final output $\mathcal{F}_{g} \in \mathbb{R}^{D \times H \times W}$. The entire sequence of operations can be formally expressed as:
\begin{equation}
\mathcal{F}_g = \tau\left(\alpha \cdot \left(\textbf{C} + \varphi\left(\textbf{M} \odot \textbf{g}\right)\right)\right),
\end{equation}
where $\odot$ denotes element-wise multiplication, and $\varphi(\cdot)$ represents a learnable transformation function that further processes the globally-informed features.

\begin{figure*}[t]
    \centering
    \begin{subfigure}[t]{0.165\linewidth}
        \centering
        \includegraphics[width=\linewidth]{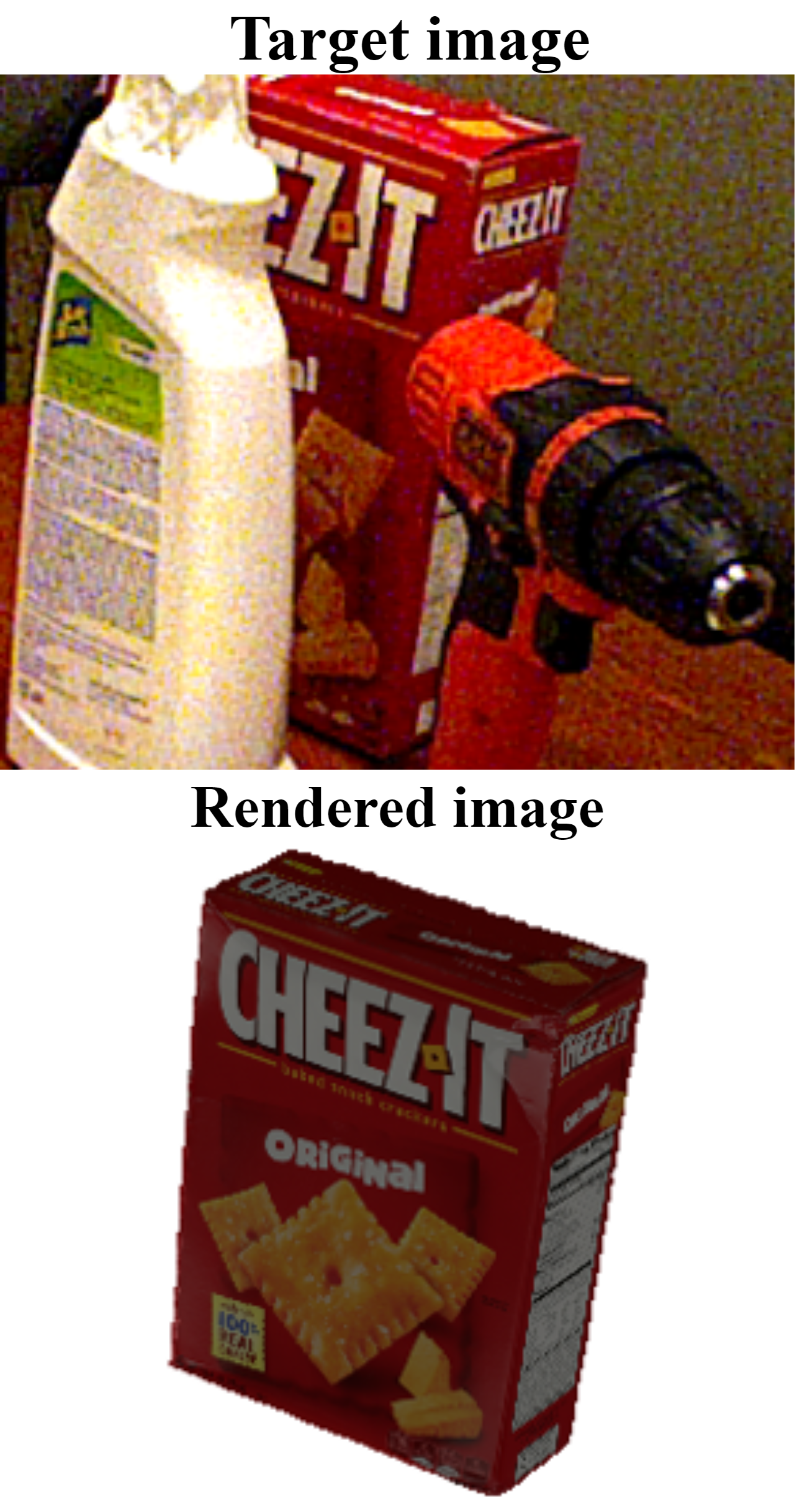}
        \caption{The target image and rendered image.}
        \label{fig: flow_error_a}
    \end{subfigure}
    \hfill
    \begin{subfigure}[t]{0.825\linewidth}
        \centering
        \includegraphics[width=\linewidth]{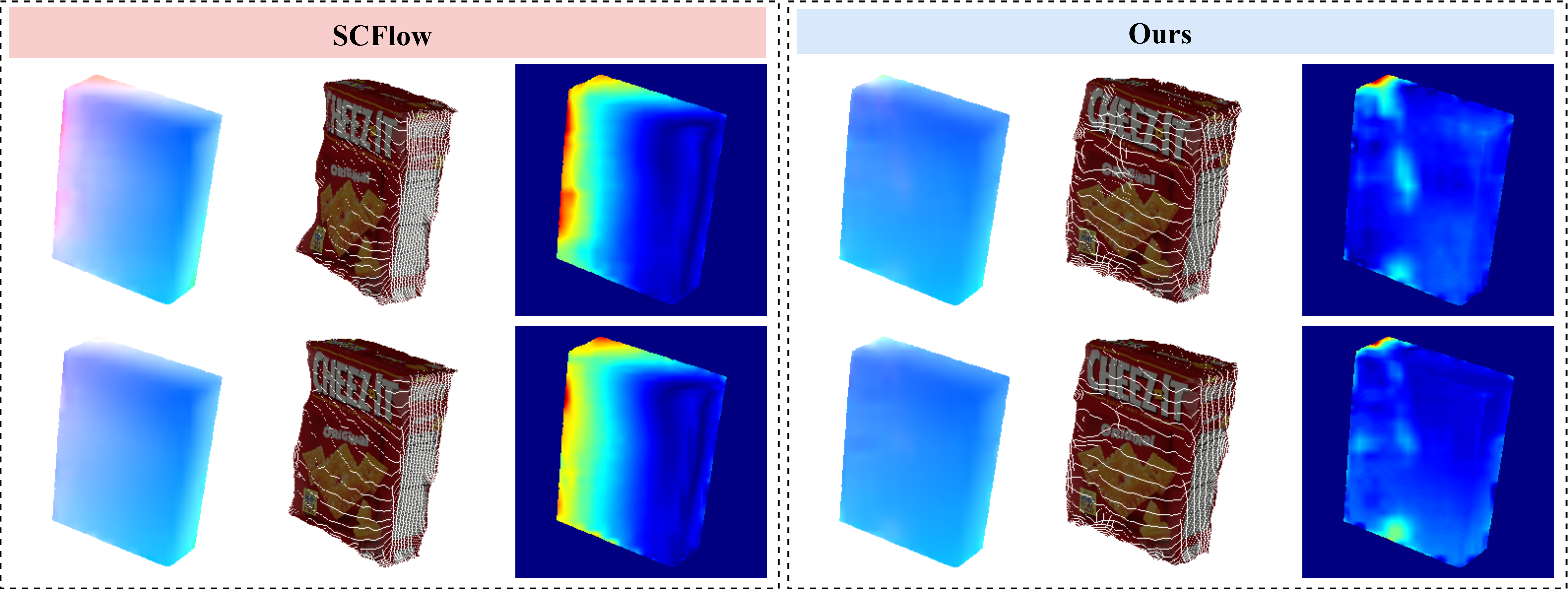}
        \caption{Results of the 2nd and 8th iterations are shown in two rows, with each row containing flow, target image reconstruction, and error heatmap.}
        \label{fig: flow_error_b}
    \end{subfigure}
    \caption{\textbf{Intermediate Flow Comparison.} Our method demonstrates superior handling of occlusions and better utilization of contextual information in early iterations.}
    \label{fig: flow_error_compare}
\end{figure*}

Now, we can concatenate the context feature $\mathcal{F}_{c}$, local motion feature $\mathcal{F}_{m}$, and global motion feature $\mathcal{F}_{g}$ as inputs to the GRU. Fig. \ref{fig: flow_error_compare} demonstrates the impact of GMC on subsequent intermediate  flow estimation. The results demonstrate that in the intermediate flow estimation task, our method exhibits rapid convergence: even in the early stages of iteration, the reconstructed target image can effectively restore the real shape. This accurate shape reconstruction capability directly results in lower  flow errors, and this advantage persists throughout the entire iterative process.

\subsection{Iterative Pose Update Operator}
\label{sec: gru}
\textbf{Flow Regression.} During the iterative stage, our method begins by using an initial  flow $f\left( u,v \right)$  to map each pixel $\boldsymbol{x}=\left( u,v \right)$ from rendered image $I_{r}$ to its corresponding position $\boldsymbol{x}^{\prime}=\left( u,v \right) +f\left( u,v \right) $ in target image $I_{t}$. Around this mapped position, we define a neighborhood $\mathcal{N} \left( \boldsymbol{x}^{\prime} \right) =\{ \boldsymbol{x}^{\prime}+\boldsymbol{dx}\left| \boldsymbol{dx}\in \mathbb{Z} ^2,\left| \boldsymbol{dx} \right|\leqslant r \right. \}$ with a radius of 4 pixels. Using this neighborhood, we extract motion features from the correlation pyramid, interpolating through bilinear sampling.
Subsequently, we input a combination of features into the GRU network: the hidden state features $h_{k-1}$ obtained in Sec. \ref{sec: extractor}, contextual information, the current  flow, and the local motion and global motion features. The GRU network thus predicts a correction value $\Delta f$ for the  flow $f\left( u,v \right)$ based on these inputs\cite{cho2014properties}. By combining the initial  flow with this correction value, we obtain an updated  flow prediction $f_{inter}=f+\Delta f$\cite{hai2023shape}.

\textbf{Flow to Pose.} Based on the flow, we can obtain the 2D-to-2D correspondences between the target image $I_{t}$ and the rendered image $I_{r}$. Since we have a clear 3D-to-2D correspondence when rendering the image, we could theoretically use the PnP algorithm to estimate the pose. However, the PnP algorithm is not stably differentiable during training. Therefore, in practice, we learn the pose correction value $\Delta \textbf{P}$ from the intermediate flow $f_{inter}$, resulting in an updated pose ${\textbf{P}_k}$.

\textbf{Pose to Flow.} Unlike traditional methodsn\cite{hu2022perspective,hai2023shape,xu2022rnnpose}, we do not directly use the predicted  flow $f_{inter}$ as the query input for the next iteratio. In the context of 6D pose estimation, this approach might lead to an unnecessarily large matching space. Instead, we project the 3D points $\textbf{p}_i$ of the target object onto the 2D plane using both the corrected pose ${\textbf{P}_k}$ and the initial pose $\textbf{P}_{0}$. Subsequently, we calculate the displacement between these two sets of projected points $\boldsymbol{\textbf{u}}_{0,i}$ and $\boldsymbol{\textbf{u}}_{k,i}$, yielding a flow $f_k$ constrained by the object's 3D shape. This process can be formulated as:
\begin{equation}
\lambda _i\left[ \begin{array}{c}
	\mathbf{u}_{j,i}\\
	1\\
\end{array} \right] =\mathbf{K}\left[ \begin{matrix}
	\boldsymbol{R}_j&		\boldsymbol{t}_j\\
	0&		1\\
\end{matrix} \right] \left[ \begin{array}{c}
	\mathbf{p}_i\\
	1\\
\end{array} \right] ,\ j\in \left\{ 0,k \right\},
\end{equation}

\begin{equation}
f_{k}=\boldsymbol{\textbf{u}}_{k,i}-\boldsymbol{\textbf{u}}_{0,i},
\end{equation}
where $\lambda _i$ is a scale factor, and $\boldsymbol{\textbf{K}}$ is the camera intrinsic matrix.

We use this newly computed  flow for feature extraction in the correlation pyramid in the next iteration. This marks the completion of one full iteration.

\subsection{Loss Function}
\label{sec: loss}
Our loss function design revolves around two core tasks:  flow estimation and pose estimation, aiming to effectively guide model optimization. For flow estimation, we adopt the mature approach from RAFT. Specifically, we calculate the ground truth flow using initial and actual pose, and compare it with the flow generated by the flow regressor. To quantify prediction errors, we use the L1 loss. Considering that the rendered images only contain the target object, we precisely limit the loss calculation to pixels within the rendered image mask.
For pose estimation, we employ the widely recognized point-matching loss. Our method involves randomly selecting one thousand representative points from the object's 3D model and applying rotation and translation transformations to these points using both the true pose and the estimated pose. By calculating the distance between these two transformed point sets, we accurately assess the precision of pose estimation.
To enhance the model's iterative learning capability, we implement a weighting mechanism: prediction weights increase exponentially with the number of iterations, causing the model to focus more on predictions from later iterations. Furthermore, given that pose estimation is our primary task, we introduce a balancing factor $k$ to adjust the relative importance of pose estimation and  flow estimation tasks. The comprehensive loss function is formulated as:
\begin{equation}
\mathcal{L} = \sum_{j=1}^N \gamma^{N-j} \left(\mathcal{L}_{\text{pose}} + k \cdot \mathcal{L}_{\text{flow}}\right),
\end{equation}
where we set the total number of iterations $N$ to 4 and 8, the parameter $\gamma$ to 0.8, and $k$ to 0.1.

\section{EXPERIMENTS}
In this section, we conduct a systematic evaluation of our proposed method. We begin by outlining the experimental setup, detailing the datasets utilized, and specifying the evaluation metrics. Subsequently, we compare our approach against state-of-the-art methods to demonstrate its effectiveness. We then perform ablation studies to validate the efficacy of individual components within our method. Finally, we provide a comprehensive analysis of the runtime performance.

\subsection{Experimental Setup}
\label{sec: setup}
\textbf{Implementation Details.} Our method employs a render-and-compare strategy \cite{iwase2021repose,li2018deepim,xu2022rnnpose,hai2023shape}. Using the Pytorch3D\cite{ravi2020pytorch3d} framework, we generate rendered image $I_r$ based on the initial pose and set its resolution to $256\times256$ pixels. For the target image $I_t$, we crop it according to the bounding box of the target object to match the resolution of the rendered image, enabling direct comparison. During training, we use the AdamW optimizer to train the model with the following configuration: an initial learning rate of 4e-4, $\beta_1$ and $\beta_2$ set to 0.9 and 0.999 respectively, an $\varepsilon$ value of 1e-8, and a weight decay coefficient of 1e-4. The batch size is set to 16. To stabilize the training process, we apply gradient clipping in the optimizer, limiting the gradient norm to 10.0.
For learning rate scheduling, we adopt the OneCycle strategy\cite{smith2019super}. The maximum learning rate is maintained at 4e-4, with a total of 200 thousand training steps. The learning rate rapidly increases to its peak value during the initial 5$\%$ of the training period, followed by a gradual decrease through linear annealing. This strategy aims to accelerate model convergence and enhance overall performance.

\begin{table*}[t]
\centering
\caption{\textbf{Comparison with the state-of-the-art methods.} Here we report the Average Recall (\%) of ADD-0.1d. N represents the number of iterations.}
\label{table: sota_compare}
\begin{tabular}{lccccccccccc}
\hline
\multirow{2}{*}{Dataset} & PoseCNN & PVNet & SO-Pose & DeepIM&GDR-Net&RePose&RNNPose&PFA&SCFlow&Ours&\textbf{Ours} \\
                  & \cite{xiangposecnn} & \cite{peng2019pvnet} & \cite{di2021so} & \cite{li2018deepim}&\cite{wang2021gdr}&\cite{iwase2021repose}&\cite{xu2022rnnpose}&\cite{hu2022perspective}&\cite{hai2023shape}& (N=4)&  \textbf{(N=8)}                \\ \hline
 LM-O&  24.9& 40.8 & 62.3 & 55.5 & 62.2 & 51.6 & 60.7 & 64.1 & 66.4 & \underline{66.6}& \textbf{67.0}\\
 YCB-V&  21.3&  -& 56.8 & 53.6 & 60.1 & 62.1 & 66.4 &62.8  &  70.5& \underline{73.3} &\textbf{74.2}\\ \hline
\end{tabular}
\end{table*}

\begin{table}[t]
\centering
\caption{Refinement comparison measured by BOP metrics.}
\label{table: bop_metric_compare}
\begin{tabular}{lcccc}

 \hline
 Method  &Avg.    & MSPD   &VSD   &MSSD   \\ \hline

\multicolumn{5}{c}{LM-O} \\ \hline
  CosyPose\cite{labbe2020cosypose} &  0.633  &0.812    &0.480   &0.606   \\
    SurfEmb\cite{haugaard2022surfemb} &   0.647 &\textbf{0.851}    &0.497   &0.640   \\
  CIR\cite{lipson2022coupled} &  0.655  & 0.831   &0.501   &0.633   \\
     PFA\cite{hu2022perspective}&   0.674 & 0.818   &0.531   &0.673   \\
      SCFlow\cite{hai2023shape}& 0.682   &0.842    &0.532  &0.674   \\
  Ours(N=4) &  \underline{0.683}  &  0.842  & \underline{0.533 } & \underline{0.676}  \\ 
\textbf{Ours(N=8)} &  \textbf{0.689}  &  \underline{0.846}  & \textbf{0.540 } & \textbf{0.681} \\ \hline
\multicolumn{5}{c}{YCB-V} \\ \hline
 SurfEmb\cite{haugaard2022surfemb}  & 0.781   & -   & -  &-   \\
   SC6D\cite{cai2022sc6d} &   0.788 & -   & -  & -  \\
      PFA\cite{hu2022perspective}&  0.795  & 0.844   &0.743   &0.797   \\
CosyPose\cite{labbe2020cosypose} &  0.821  & 0.850   &0.772   & \underline{0.842}  \\
  CIR\cite{lipson2022coupled} &  0.824  &0.852    &0.783   &0.835   \\
   SCFlow\cite{hai2023shape}&   0.826 & 0.860   & 0.772  & 0.840  \\
  Ours(N=4) &  \underline{0.831}  & \underline{0.866}   &  \underline{0.785} &  \underline{0.842} \\ 
\textbf{Ours(N=8)} &  \textbf{0.835}  & \textbf{0.869}   &  \textbf{0.788} &  \textbf{0.849} \\ \hline

\end{tabular}
\end{table}
\textbf{Datasets.}
We evaluated our method on two common benchmark datasets: LM-O\cite{cho2014properties} and YCB-V\cite{xiangposecnn}. The LM-O dataset contains 8 target objects that frequently overlap or are partially occluded in images, increasing the difficulty of pose estimation. While many methods train separate networks for each object in LM-O, we trained a single network to handle all objects for practicality and efficiency. The YCB-V dataset contains 21 everyday objects with diverse shapes, sizes, and materials, and includes challenges such as occlusion, lighting variations, and complex backgrounds. Similarly, we trained a single network to handle all 21 objects rather than training separate networks for each object. For both datasets, we additionally use physically-based rendering (PBR) data during training.

\textbf{Evaluation Metrics.} This study employs the widely used ADD and ADD-S precision evaluation metrics in the field of 6D pose estimation. ADD (Average Distance of Model Points) applies to asymmetric objects, calculating the average Euclidean distance between object surface points under estimated and ground truth poses. ADD-S  is used for symmetric objects, computing the average distance between each point in the estimated pose and its nearest point in the ground truth pose.
We utilize specific metrics such as ADD-0.05d and ADD-0.1d, where d represents the object diameter. An estimation is considered correct when the average estimated distance is less than 5$\%$ or 10$\%$ of the object's diameter. These thresholds are used to assess the accuracy of pose estimation. 
To compare with the latest research, we also compute three standard metrics from the BOP benchmarks:
Visible Surface Discrepancy (VSD), Maximum Symmetry-aware Surface Distance (MSSD) and Maximum Symmetry-aware Projection Distance (MSPD).
These metrics have become commonly reported indicators in recent studies. Readers can refer to \cite{hodavn2020bop} for detailed definitions of these metrics.
\subsection{Comparison with State of the Art}

We compare our proposed method with current state-of-the-art techniques, including PoseCNN\cite{xiangposecnn}, PVNet\cite{peng2019pvnet}, SO-Pose\cite{di2021so}, DeepIM\cite{li2018deepim}, GDR-Net\cite{wang2021gdr}, RePose\cite{iwase2021repose}, RNNPose\cite{xu2022rnnpose}, PFA\cite{hu2022perspective}, and SCFlow\cite{hai2023shape}. To ensure a fair comparison, we adopt the same strategy as other methods, using PoseCNN for pose initialization. Tab. \ref{table: sota_compare} presents detailed comparison results, demonstrating that our method achieves the best performance. As can be seen, our method outperforms other methods in both cases when the number of iterations is 4 and 8.

\begin{figure*}[t]
    \centering
    \includegraphics[width=\linewidth]{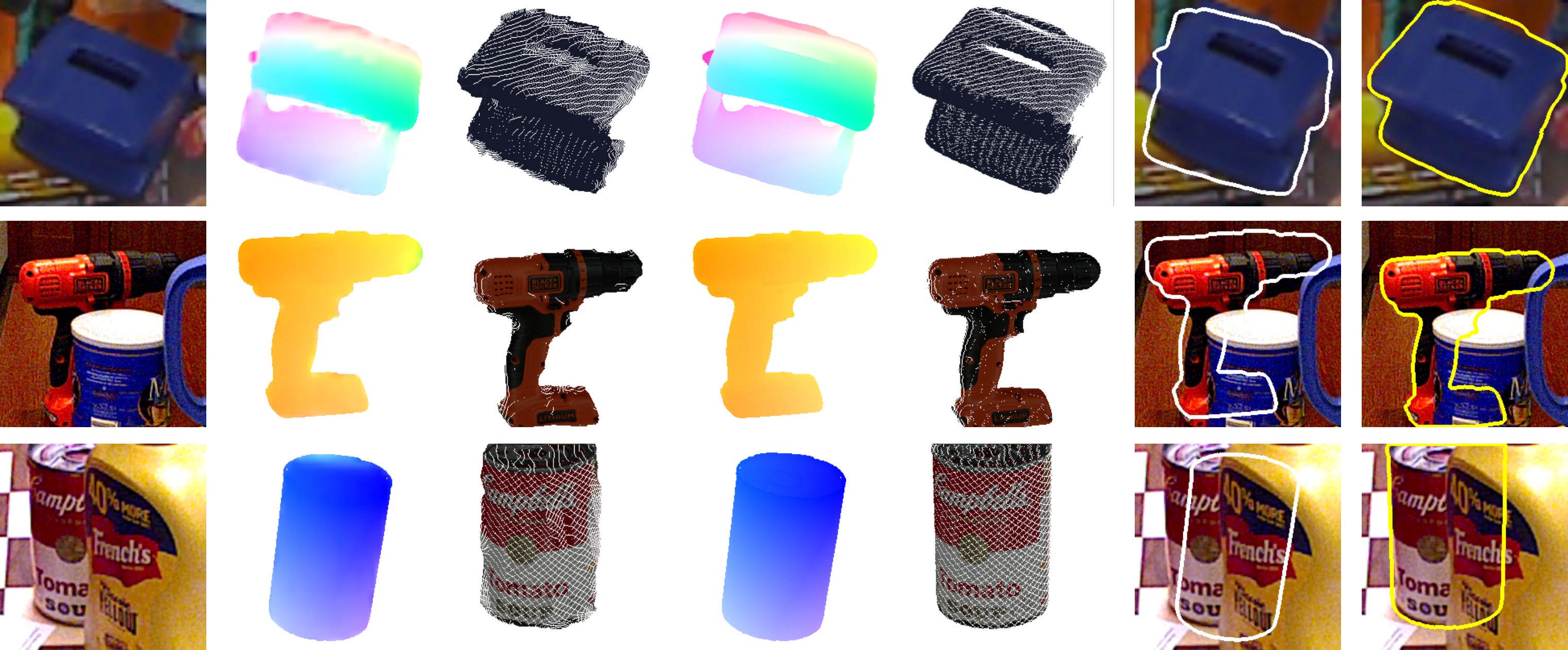}
    \caption{\textbf{Qualitative  results.} From left to right: target image, intermediate  flow with reconstruction, pose-induced flow with reconstruction, and comparison before and after our method. Three rows illustrate cases where the target object is complete, occluded, and partially cropped, respectively.}
    \label{fig: result_show}
\end{figure*}

Furthermore, we carry out additional comparisons with methods that report BOP benchmark metrics, such as CIR\cite{lipson2022coupled}, CosyPose\cite{labbe2020cosypose}, SurfEmb\cite{haugaard2022surfemb}, and SC6D\cite{cai2022sc6d}. Again, to maintain fairness, we follow the approach of these methods by using CosyPose for pose initialization. Tab. \ref{table: bop_metric_compare} shows the detailed comparison results under the BOP metrics. The results indicate that our method also exhibits superior performance under these metrics.
Fig. \ref{fig: result_show} visually demonstrates the effectiveness of our method. Our method effectively refines the initial pose estimation with high accuracy, regardless of whether the target object is fully visible in the image.

\begin{figure}[t]
    \centering
    \begin{subfigure}[t]{0.49\linewidth}
        \centering
        \includegraphics[width=\linewidth]{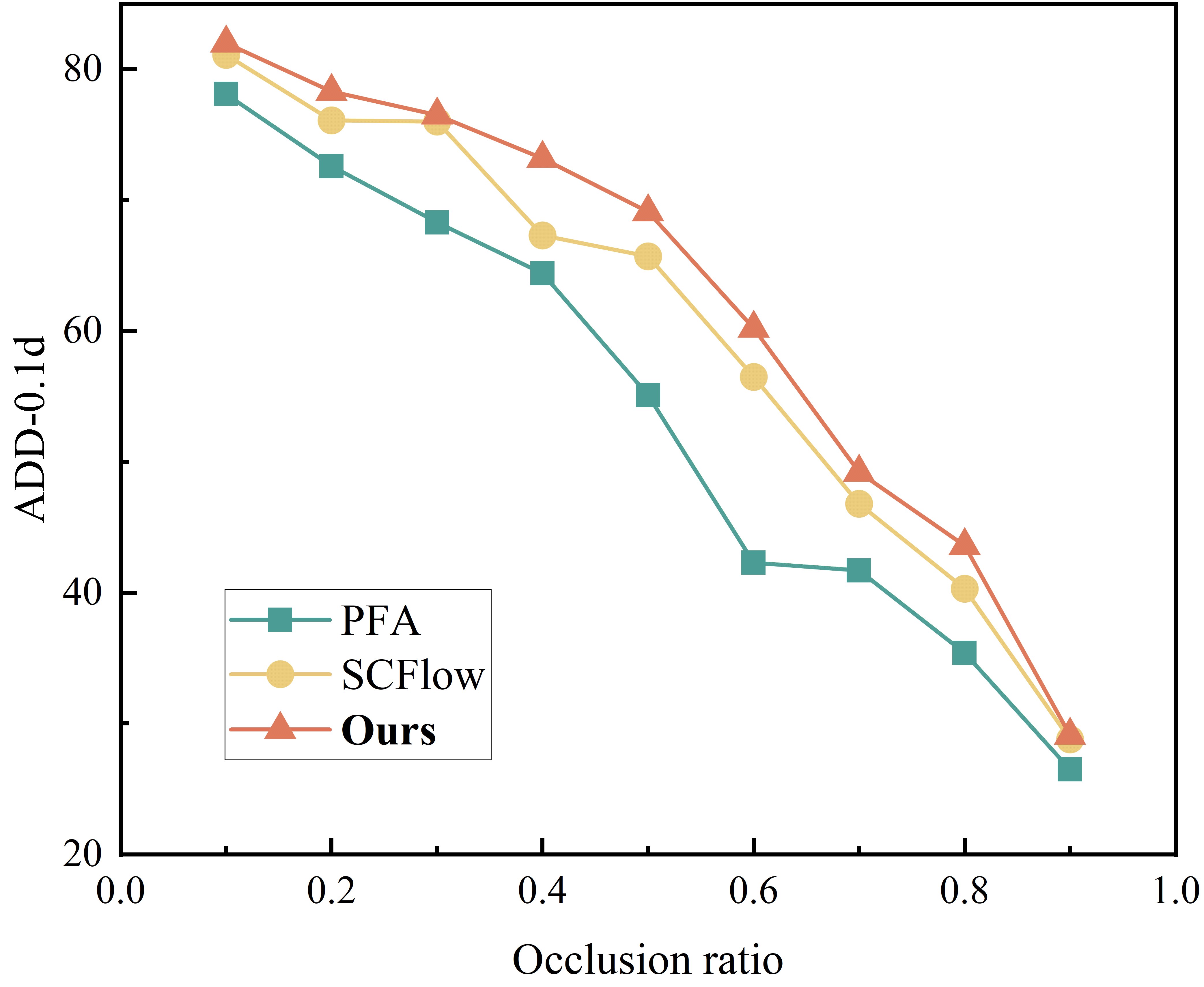}
        \caption{Results on LM-O.}
        \label{fig: occ_lmo}
    \end{subfigure}
    \hfill
    \begin{subfigure}[t]{0.49\linewidth}
        \centering
        \includegraphics[width=\linewidth]{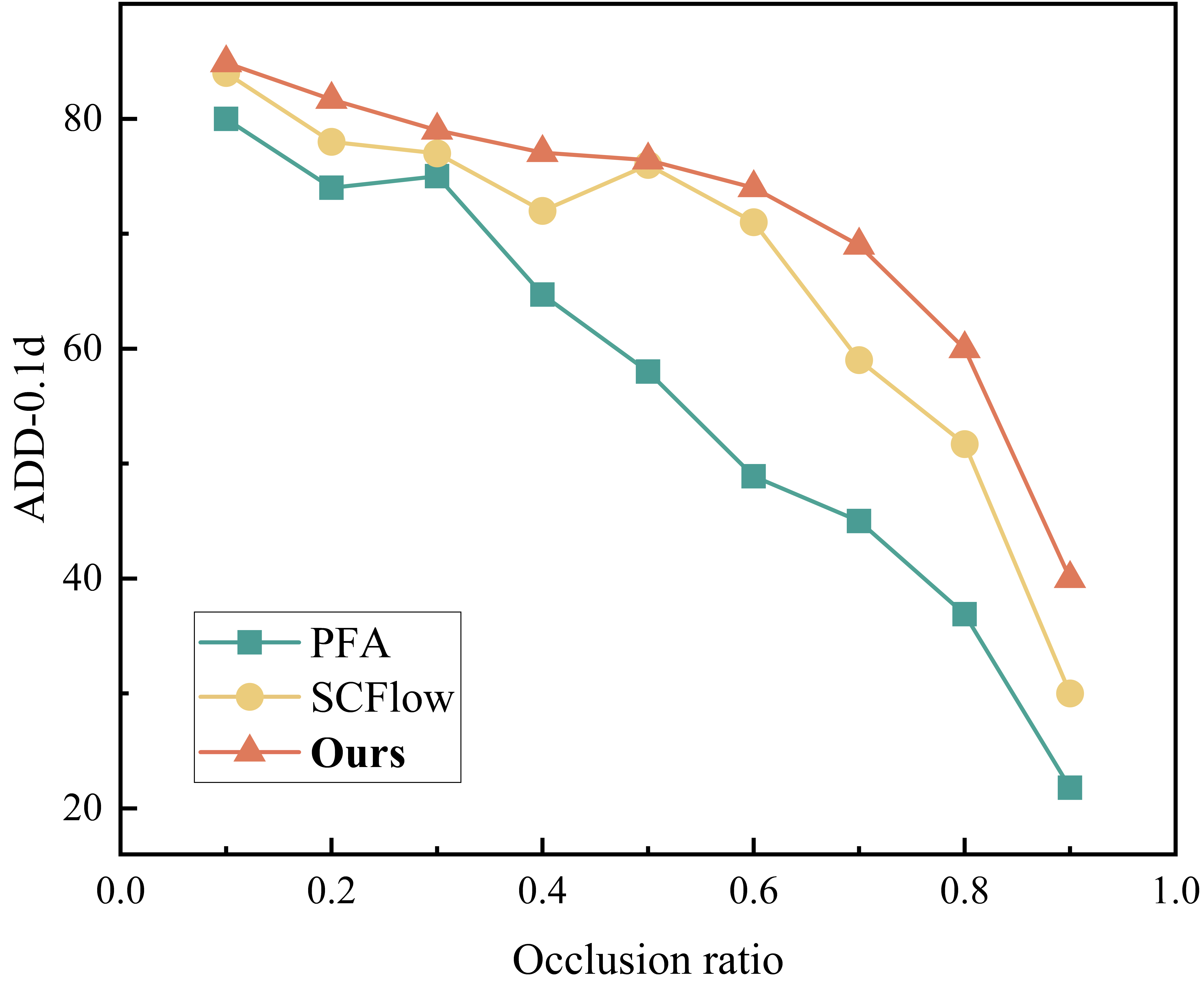}
        \caption{Results on YCB-V.}
        \label{fig: occ_ycbv}
    \end{subfigure}
\caption{Results under different degrees of occlusion.}
    \label{fig: occ_compare}
\end{figure}

\begin{figure}[t]
    \centering
    \begin{subfigure}[t]{0.49\linewidth}
        \centering
        \includegraphics[width=\linewidth]{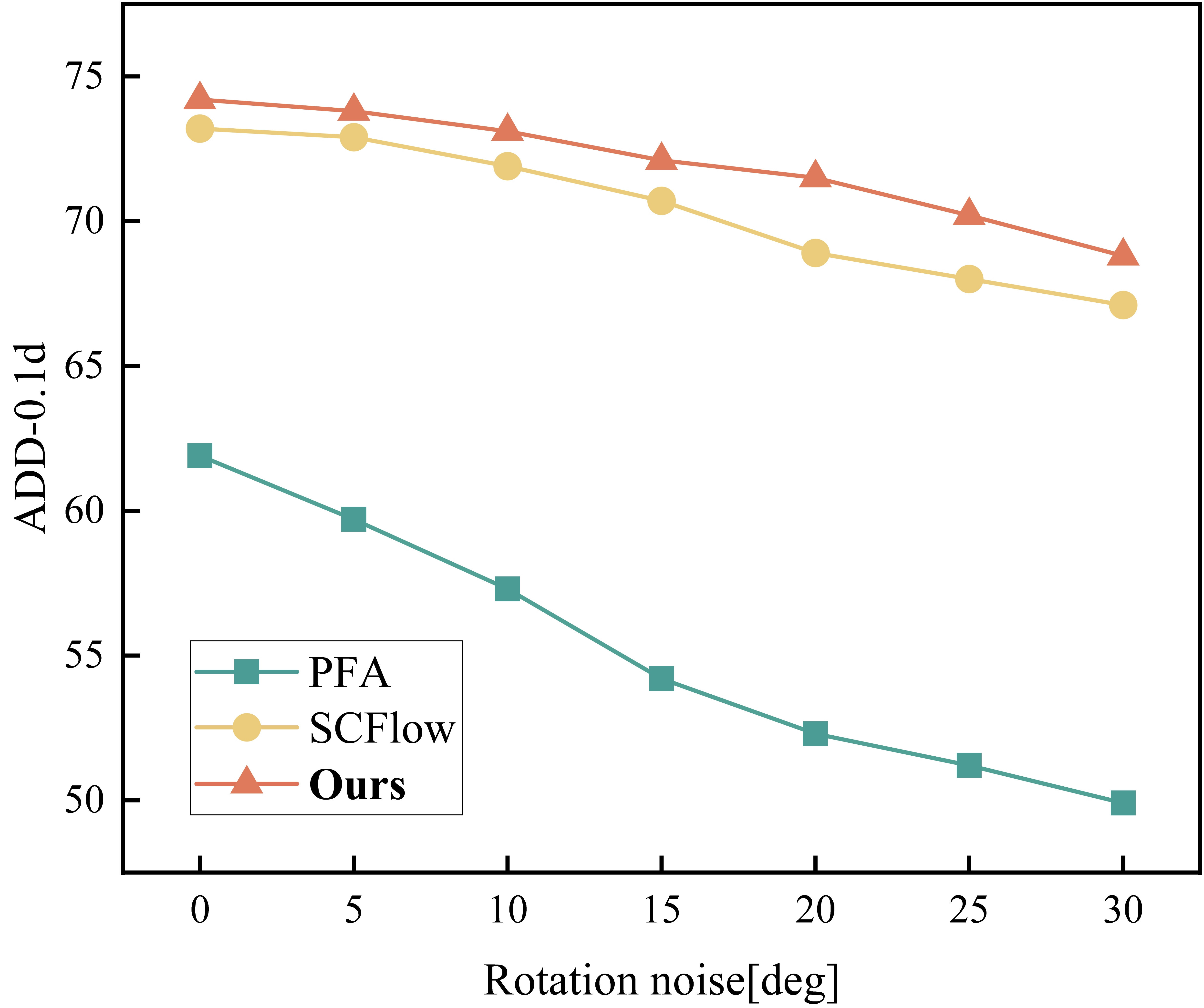}
        \caption{Results across varying angular noise levels.}
        \label{fig: noise}
    \end{subfigure}
    \hfill
    \begin{subfigure}[t]{0.49\linewidth}
        \centering
        \includegraphics[width=\linewidth]{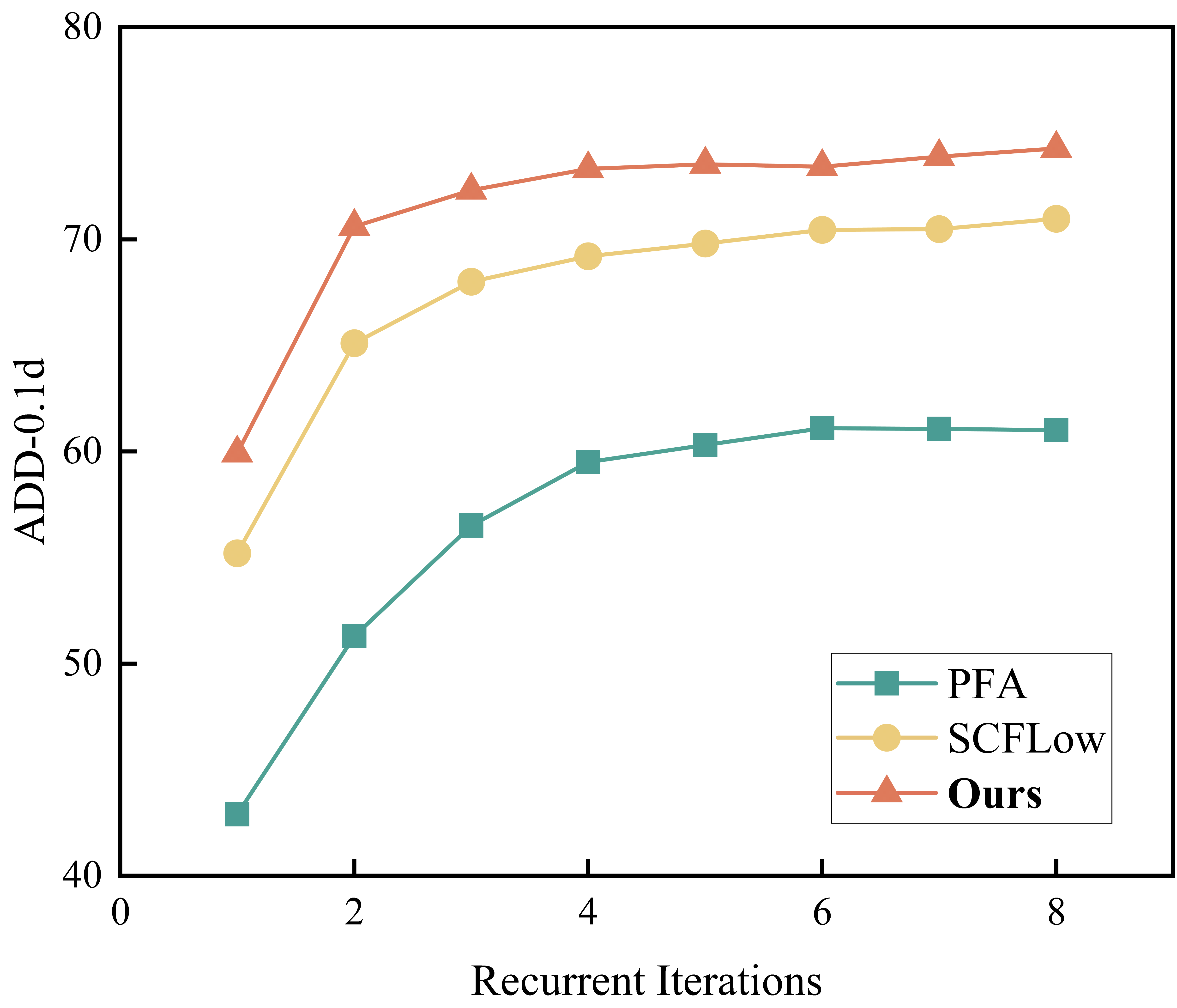}
        \caption{Results with different recurrent iterations during inference.}
        \label{fig: iteration}
    \end{subfigure}
\caption{Comparison of robustness and efficiency.}
    \label{fig: robustness and efficiency}
\end{figure}

\begin{table}[t]
\caption{Ablation Study on YCB-V.}
\label{table: ablation}
\begin{tabular}{l|l|cc}
\hline
\multirow{2}{*}{Row} & \multirow{2}{*}{Method} & ADD  & ADD  \\
                  &                   & 0.05d & 0.1d \\ \hline
                \textbf{A0}  & \textbf{GMFlow (ours)}                  &  \textbf{55.6} & \textbf{74.2} \\
                  B0&  A0 $\rightarrow$ No global motion info. aggregation                 & 53.3 & 72.1 \\
                 C0& A0 $\rightarrow$ No shape constraints on the flow                   & 48.6 & 69.6   \\
                 D0& WDR results\cite{hu2021wide}               & 5.2 & 27.5 \\ 
                  D1& A0: Initialization method PoseCNN $\rightarrow$ WDR               &48.1 & 69.4  \\
                   D2  &   A0: Adds angular noise (std. dev. 25°)      & 50.5 & 70.2   \\
                 E0 &A0: Iterations 8 $\rightarrow$ 4                  & 53.9 & 73.3\\ \hline
            
\end{tabular}
\end{table}
\subsection{Ablation Study}
We conduct several ablation experiments on the YCB-V dataset, and the results are shown in Tab. \ref{table: ablation}. All results are for the same set described in Sec. \ref{sec: setup}.

\textbf{Global information aggregation.} Global motion information aggregation is the core module designed to address occlusion and information loss challenges in 6D pose estimation scenarios using flow methods. In row B0 of Tab. \ref{table: ablation}, we demonstrate the performance of our method without this module.
At error thresholds of 0.05 and 0.1 times the grid diameter, our average recall decreases by 2.3 and 2.1 percentage points, respectively.
To provide a more comprehensive and intuitive assessment of our method's capability to handle occlusions, we conduct an extensive experimental evaluation across occlusion ratios ranging from 0.1 to 0.9. As shown in Fig. \ref{fig: occ_compare},  our approach consistently outperforms both PFA and SCFlow across the entire spectrum of occlusion levels.

\textbf{Shape constraints to the flow from pose.}
The global motion information introduces structural constraints for the intermediate flow. Correspondingly, we also introduce shape constraints for the flow of pose transformation. In the C0 row, we remove this constraint, leading to a decrease in average recall rates. We believe that both the intermediate flow and the pose-transformed flow require structural constraints. This approach allows us to fully leverage the information that the target object for pose estimation is a rigid body.

\textbf{Different pose initialization.} To verify the generalizability and robustness of our method, we conducted comprehensive experimental validation. First, we demonstrated that our approach is not limited to specific initialization results, but can flexibly integrate with various pose initialization methods.
In the experiments, we performed inference analysis on different initial pose values. Specifically, the D0 row displays the original pose estimation results based on WDR\cite{hu2021wide}, while the D1 row presents the average recall after further optimization on the WDR initialization. The experimental results clearly show that our method significantly improved the accuracy of initial pose estimation.

To more comprehensively evaluate the method's robustness, we conducted more challenging experiments. We introduced angular noise of varying degrees along the three axes of pose initialization, with the standard deviation progressively increasing from 5 to 30 degrees. As shown in Fig. \ref{fig: robustness and efficiency}(a), even under such significant noise interference, our method still demonstrates exceptional pose optimization capabilities, consistently outperforming comparative methods like PFA and SCFlow.

\textbf{Different number of recurrent iterations.} In Fig. \ref{fig: robustness and efficiency}(b), we systematically presented experimental results across 1 to 8 iterations. A notable finding is that our method, after just 2 iterations, can already approach the performance level of SCFlow after 8 iterations, which means we can significantly reduce computational overhead.
This performance advantage can be attributed to our designed global motion information aggregation strategy. Through this approach, our internal optical flow estimation can accurately capture and reconstruct structural features in the early stages of iteration, making the results closer to the true values. This is consistent with the visualization results in Fig. \ref{fig: flow_error_compare}, intuitively validating the effectiveness of our method.

\subsection{Runtime Analysis}
\begin{table*}[]
\centering
\caption{\textbf{Runtime comparison.} We report runtime comparisons for processing images containing a single object instance.}
\label{table: time_compare}
\begin{tabular}{lcccccccc}
\hline
 Method& CIR\cite{lipson2022coupled} & SurfEmb\cite{haugaard2022surfemb}  & CosyPose\cite{labbe2020cosypose}&DeepIm\cite{li2018deepim} & PFA \cite{hu2022perspective}& SCFlow\cite{hai2023shape} & \textbf{Ours(N=4)} & Ours(N=8)\\ \hline
 Timeing [ms]&11k & 1k  & 110& 82 & 38 & \underline{17} & \textbf{13} & 18\\ \hline
\end{tabular}
\end{table*}
DeepIM\cite{li2018deepim}, CosyPose\cite{labbe2020cosypose}, and CIR\cite{lipson2022coupled} employ multiple rendering and iteration steps during their inference process, resulting in prolonged execution times. SurfEmb, in its pose refinement phase, requires a dense and continuous distribution of 2D-3D correspondences, which incurs significant time costs.
More recent approaches like PFA\cite{hu2022perspective} and SCFlow\cite{hai2023shape} utilize dense 2D correspondence fields between rendered images of coarse pose estimates and real images to infer necessary pose corrections. While this approach indeed reduces runtime, PFA's execution time remains constrained by multiple iterative rendering steps.
In contrast, our method offers significant improvements in both efficiency and accuracy. It requires only one rendering step during inference and achieves higher-quality flow estimation in each iteration. Notably, while SCFlow requires 8 iterations, our method achieves higher accuracy with only 4 iterations while maintaining shorter runtime. The specific runtime comparisons are presented in Tab. \ref{table: time_compare}.

\section{CONCLUSIONS}
This letter presents GMFlow, a global motion-guided recurrent flow estimation method for 6D object pose estimation. 
Existing refinement methods struggle to effectively address local ambiguities caused by occlusion or partial object absence. In contrast, GMFlow overcomes this challenge through global interpretation. It processes global contextual features using a linear attention mechanism, which then guides the local motion  to produce comprehensive global motion estimates. Additionally, we introduce object shape constraints in the flow iteration process to make flow estimation more suitable for pose estimation scenarios. Experiments on the LM-O and YCB-V datasets demonstrate that our method outperforms existing techniques in accuracy while maintaining competitive computational efficiency.




\end{document}